 \documentclass[smallabstract,smallcaptions]{dccpaper}

\usepackage{epsfig}
\usepackage{amsmath}
\usepackage{amssymb}
\usepackage{color}
\usepackage{url}
\usepackage{booktabs}
\usepackage{subfigure}
\usepackage{multirow}
\newlength{\figurewidth}
\newlength{\smallfigurewidth}

\begin{document}

\title
{\large
\textbf{UCVC: A Unified Contextual Video Compression Framework with Joint P-frame and B-frame Coding}
}

\author{%
%Author One$^{\ast}$, Author Two$^{\dag}$, and Author Three$^{\ast}$\\[0.5em]
Jiayu Yang$^{1}$, Wei Jiang$^{1}$, Yongqi Zhai$^{1,2}$, Chunhui Yang$^{1}$, Ronggang Wang$^{1,2,3}$\\[0.5em]
{\small\begin{minipage}{\linewidth}\begin{center}
\begin{tabular}{c}
$^{1}$School of Electronic and Computer Engineering, Peking University, China\\
$^{2}$Peng Cheng Laboratory, China \\
$^{3}$Migu Culture Technology Co., Ltd, China \\
\url{jiayuyang@pku.edu.cn, rgwang@pkusz.edu.cn} 
\end{tabular}
\end{center}\end{minipage}}
\thanks{This work is financially supported by National Natural Science Foundation of China U21B2012 and  62072013, Shenzhen Science and Technology Program-Shenzhen Cultivation of Excellent Scientific and Technological Innovation Talents project(Grant No. RCJC20200714114435057) , Shenzhen Science and Technology Program-Shenzhen Hong Kong joint funding project (Grant No. SGDX20211123144400001), this work is also financially supported for Outstanding Talents Training Fund in Shenzhen. Corresponding author: Ronggang Wang.}
}

\maketitle
\thispagestyle{empty}

\begin{abstract}
This paper presents a learned video compression method in response to video compression track of the 6th Challenge on Learned Image Compression (CLIC), at DCC 2024.
Specifically, we propose a unified contextual video compression framework (UCVC) for joint P-frame and B-frame coding. Each non-intra frame refers to two neighboring decoded frames, which can be either both from the past for P-frame compression, or one from the past and one from the future for B-frame compression. 
In training stage, the model parameters are jointly optimized with both P-frames and B-frames. 
Benefiting from the designs, the framework can support both P-frame and B-frame coding and achieve comparable compression efficiency with that specifically designed for P-frame or B-frame.
As for challenge submission, we report the optimal compression efficiency by selecting appropriate frame types for each test sequence. Our team name is PKUSZ-LVC.
%Experimental results demonstrate that our method achieves comparable compression efficiency with both traditional and recent learned video compression methods.
\end{abstract}

\section{Introduction}
Learned video compression has developed rapidly and demonstrated competitive compression efficiency compared with traditional video coding standards such as H.264 \cite{wiegand2003overview}, H.265 \cite{sullivan2012overview} and H.266 \cite{bross2021overview}. Different from traditional methods that need hand-crafted design of coding tools and can hardly be jointly optimized, learned methods \cite{lu2019dvc, hu2021fvc, li2021deep, sheng2022temporal, li2022hybrid, li2023neural, wu2018video, djelouah2019neural, yilmaz2021end, akin2023multi} can be end-to-end optimized with rate-distortion loss function to improve overall compression efficiency. 
%Recently, DCVC-DC \cite{li2023neural} reported better results than the under-developing next generation traditional codec ECM

\begin{figure}[]
\centering
\includegraphics[width=0.92\linewidth]{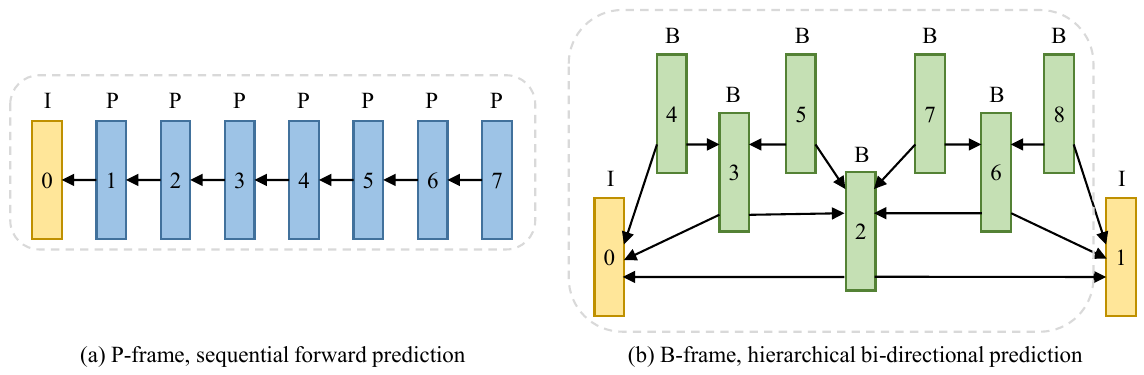} % Reduce the figure size so that it is slightly narrower than the column.
\caption{Comparison on reference structure of P-frame (left) and B-frame (right). The dashed line represents a GoP of size 8 and the numbers indicate encoding order count.}
\label{configuration}
\end{figure}

Existing learned video compression works have designed efficient compression framework to improve compression efficiency, which can be roughly divided into two categories according to  application scenarios, i.e., P-frame compression works for low latency scenario and B-frame compression works for high efficiency scenario. 
A frame that only takes past frames as references is named as predictive (P) frame while the one that takes both past frames and future frames as references is named as bi-directional predictive (B) frame, whose reference structures are shown in Fig. \ref{configuration}. The former has the advantage of low latency due to the sequential prediction, while the latter tends to achieve higher compression efficiency by referring future information.
For P-frame compression, the pionnering work DVC \cite{lu2019dvc} proposed to employ optical flow as motion representations and reduce temporal redundancy in pixel domain. Later work FVC \cite{hu2021fvc} improved prediction performance by performing inter prediction in high-dimentional feature domain to explore the potential of richer reference contexts. Recently,  DCVC series \cite{li2021deep, sheng2022temporal, li2022hybrid, li2023neural} extended residual coding to conditional coding by replacing the subtraction operation with neural networks to better utilize the predicted contexts and achieved the state-of-the-art compression efficiency.
For B-frame compression, Wu et al. \cite{wu2018video} first proposed to perform compression by bi-directional interpolation. Later work \cite{djelouah2019neural} transmitted bi-directional motions and blending coefficients to improve bi-directional prediction efficiency. LHBDC \cite{yilmaz2021end} further explored the potential of hierarchical bi-directional reference structure. Recently, LHBDC was improved with multi-scale conditional coding and reported the state-of-the-art compression efficiency on B-frame compression \cite{akin2023multi}.

Though great progress has been made, existing methods only support either P-frame or B-frame coding, which cannot flexibly choose different frame types to meet the requirements of different application scenarios. Besides, compared with compressing all sequences with a specific frame type, performing frame type selection in different motion scenarios can achieve better compression efficiency.

In response to video compression track of the 6th Challenge on Learned Image Compression (CLIC) at DCC 2024, we propose a unified contextual video compression framework (UCVC) that supports both P-frame and B-frame coding. Benefiting from the flexibility, we report the optimal compression efficiency by selecting appropriate frame type for each sequence.
In our design, each non-intra frame employs two neighboring decoded frames as references, which can be either both from the past for P-frame compression, or one from the past and one from the future for B-frame compression. 
Besides, we propose to optimize the network parameters with both P-frames and B-frames in training stage.
In this way, the framework can adapt to different frame types and achieve comparable efficiency on both two frame types with that specifically optimized with P-frames or B-frames.
Benefiting from the above design, the framework can flexibly switch to P-frame or B-frame compression.

\begin{figure*}[!t]
\centering
\includegraphics[width=\linewidth]{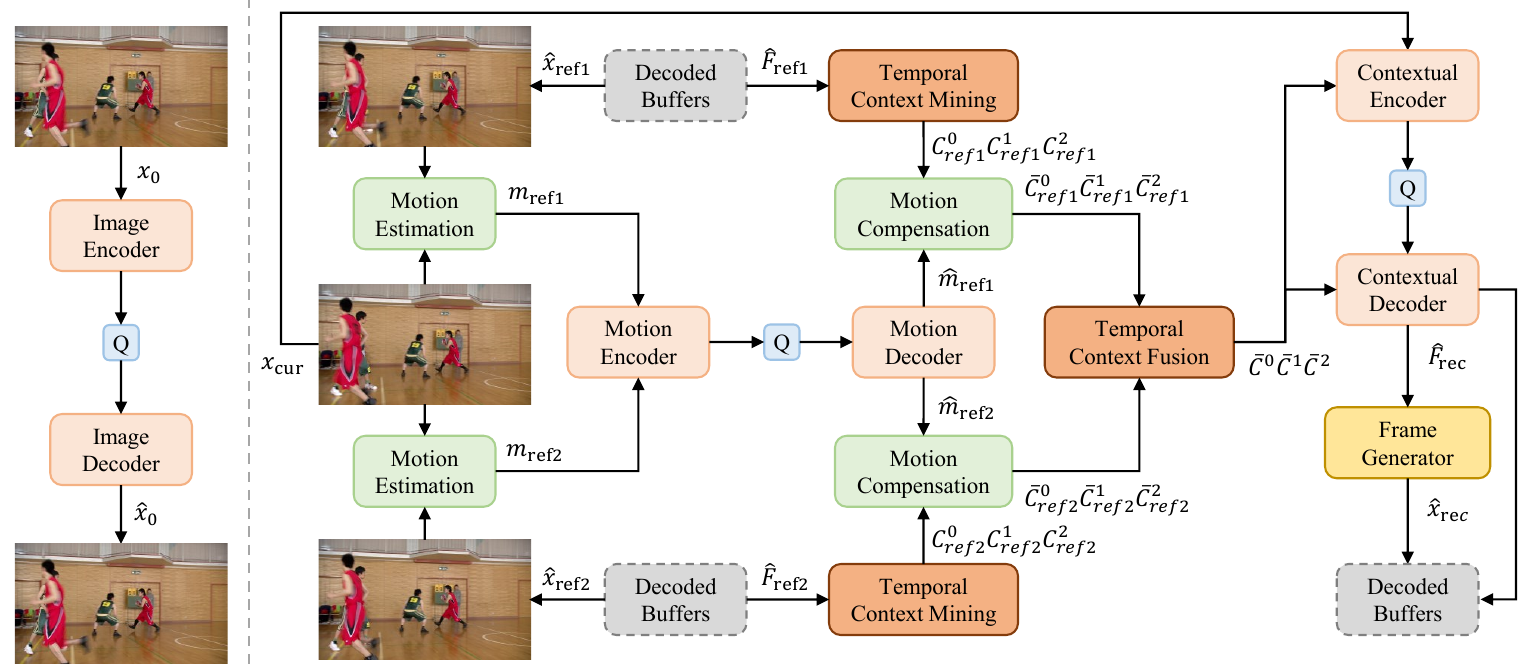} % Reduce the figure size so that it is slightly narrower than the column.
\caption{
%Due to the quality fluctuation in compressed videos, key frames can be utilized to enhance predictive frames.
Overview of our learned video compression framework. The first frame in each GoP is compressed as I-frame (left), while the others are compressed as P-frames or B-frames (right).
Given current frame $x_{cur}$ at time step $t$, the two reference frames $\hat{x}_{ref1}$ and $\hat{x}_{ref2}$ are either both from the past, e.g., $\hat{x}_{t-1}$ and $\hat{x}_{t-2}$, for P-frame compression, or one from the past and one from the future, e.g., $\hat{x}_{t-1}$ and $\hat{x}_{t+1}$, for B-frame compression.}
\label{framework}
\end{figure*}

\section{Method}
\label{Method}

\subsection{Compression Framework}
The compression framework follows conditional-coding based methods \cite{sheng2022temporal}, which consists of image compression module, motion estimation module, motion compression module, motion compensation module, temporal context mining module, temporal context fusion module, contextual compression module and frame generator module. 
The image, motion and contextual compression modules share the same network architectures but the parameters are independent. 
Specifically, we introduce \textit{cheng2020\_anchor} \cite{cheng2020image} that consists of stacked convolutional layers and residual blocks as the encoder and decoder backbones. Besides, we use scaling factors before quantization \cite{Cui_2021_CVPR} to perform variable bitrate compression.
For entropy model, we enable the mean-scale hyperprior in \cite{minnen2018joint} and disable autoregressive prior due to its complexity.

An overview of the framework is shown in Fig. \ref{framework}. Each video sequence is divided into groups of pictures (GoP) and compressed separately. The first frame in each GoP is compressed with the image compression module as Intra (I) frame, while the others are compressed with motion and contextual compression module as either forward Predictive (P) frames or bi-directional predictive (B) frames. 
Both P-frames and B-frames refers to two reference frames $\hat{x}_{ref1}$ and $\hat{x}_{ref2}$. 
For P-frame compression, the two reference frames are exactly the recent two decoded frames. For B-frame compression, the two reference frames are two nearest available frames in the hierarchical reference structure. 
Given current frame $x_{cur}$, we leverage SpyNet \cite{ranjan2017optical} to learn optical flow $m_{ref1}$, $m_{ref2}$ from two reference frames as motion information and compress the motions with motion compression module. 
Then the temporal context mining module learns multi-scale temporal contexts $C^{l}_{ref1}$, $C^{l}_{ref2}$ from the propogated features $\hat{F}_{ref1}$ and $\hat{F}_{ref2}$, where three levels are used as that in DCVC-TCM. The extracted contexts are motion compensated with decoded optical flows $\hat{m}_{ref1}$, $\hat{m}_{ref2}$ by a \textit{warp} operation and fused by concatenation along channel dimension.
Finally, the fused contexts $\Bar{C}^{l}$ are re-filled in the contextual encoder-decoder, entropy model (omitted in Fig. \ref{framework} for simplicity) and frame generator to reconstruct feature $\hat{F}_{rec}$ and frame $\hat{x}_{rec}$.
The details of temporal context mining and re-filling operations can be found in DCVC-TCM \cite{sheng2022temporal}.

\begin{figure}[]
\centering
\includegraphics[width=0.8\linewidth]{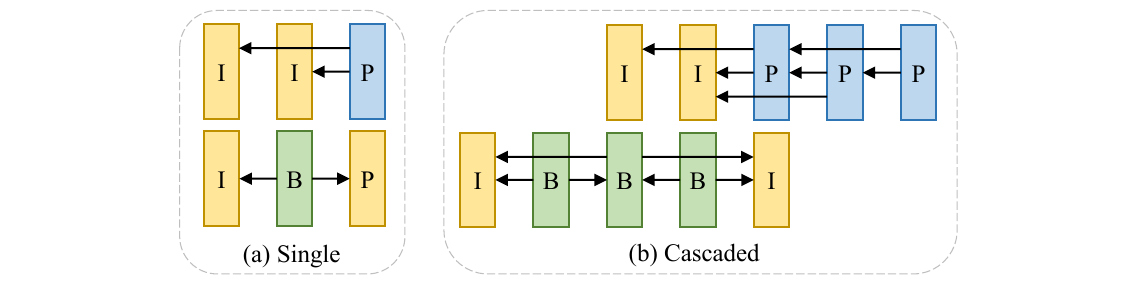} % Reduce the figure size so that it is slightly narrower than the column.
\caption{GoP structure in training stage, where each training sample contains both P-frames and B-frames.}
\label{training strategy}
\end{figure}

\subsection{Joint Training strategy}
To support both P-frame and B-frame coding, the framework should be jointly trained with both two types of frames. Specifically, each training sample contains an even number of frames, half of which are P-frames and the other half are B-frames. In this way, the framework can adapt to different frame types and achieve comparable compression efficiency with that specifically trained with one frame type.

The training strategy is shown in Fig. \ref{training strategy}, which contains two stages, i.e., \textit{Single} and \textit{Cascaded}. Following DCVC-TCM, we first use one frame for warming up and then extend to multiple frames to alleviate temporal error propagation.
In \textit{Single} stage, the frames with index 0 and 2 are compressed as I-frame, which are then used as reference frames to compress the intermediate B-frame with index 1. 
Similarly, the frames with index 0 and 1 are compressed as I-frame to be referred by the subsequent P-frame with index 2.
Then the network parameters can be optimized with both P-frame and B-frame.
In \textit{Cascaded} stage, the frames with index 0 and 4 are compressed as I-frames to be referred by the intermediate B-frames with index \{1,2,3\}.
The frames with index 2 and 3 are compressed as I-frames, which are then used as reference frames to compress the subsequent P-frames with index \{4,5,6\}.
% In this way, model parameters can be optimized with both P-frames and B-frames.
The loss function is defined as rate-distortion loss:
\begin{equation}
Loss = R + \lambda D,
\end{equation}
where $R(\cdot)$ denotes estimated entropy of encoded latents, including the side information from hyperprior, and $D(\cdot)$ represents MSE loss or MS-SSIM loss. 
% $N$ is $2$ for \textit{Single} training and $6$ for \textit{Cascaded} training.

\section{Experiments}
\label{Experiment}
\subsection{Datasets}
We use the Vimeo-90k \cite{xue2019video} training split for training. For each video sequence, we randomly crop the frames into patches with size of 256 $\times$ 256. Batch size is 8. 

We report the results on CLIC validation set following the required test conditions, i.e., 0.05 mbps and 0.5 mbps averaged on all sequences.
Besides, to compare with existing learned methods, we evaluate the performance of our method on several widely used benchmark datasets, including HEVC test sequences (Class B, C, D, E) \cite{sullivan2012overview}, UVG dataset \cite{mercat2020uvg} and MCL-JCV dataset \cite{wang2016mcl}. The HEVC test sequences contain 16 videos with different resolutions from $416 \times 240$ to $1920 \times 1080$. The UVG dataset includes seven 1080P sequences at 120 fps. And MCL-JCV dataset contains 30 videos in 1080P with diverse contents including animations.

\subsection{Implementation Details}
Our framework is implemented with PyTorch based on CompressAI \cite{begaint2020compressai} and DCVC-TCM \cite{sheng2022temporal}.
In training stage, we set $\lambda$ groups of loss function as $\{0.004, 0.01, 0.02, 0.04\}$ for PSNR models. When optimizing for MS-SSIM, we fine-tune the PSNR-optimized model with distortion loss defined as $1-\mathrm{MSSSIM}$, where $\lambda$ groups are $\{2, 6, 12, 20\}$.
For challenge submission, we optimize the model for PSNR with $\lambda$ groups as one tenth and one percent of the regular scheme to meet the requirement of target bit rate constraint.
We follow traditional compression methods and allocate more bits to I-frame by setting lambda groups of I-frame to be 5 times larger than that of P-frame and B-frame.
The model is trained with AdamW optimizer \cite{loshchilov2018decoupled} with default hyper-parameter settings. Learning rate is initialized to $1e^{-4}$ for 500k iterations, in which \textit{Cascaded} stage accounts for 400k. After that, we drop the learning rate to $1e^{-5}$ for additional 100k iterations. 
For image, motion and contextual compression modules, we set the channel number of convolutional layers as 192, 128 and 128. 

\begin{figure*}[t]
\centering
\subfigure{
\includegraphics[width=0.45\linewidth]{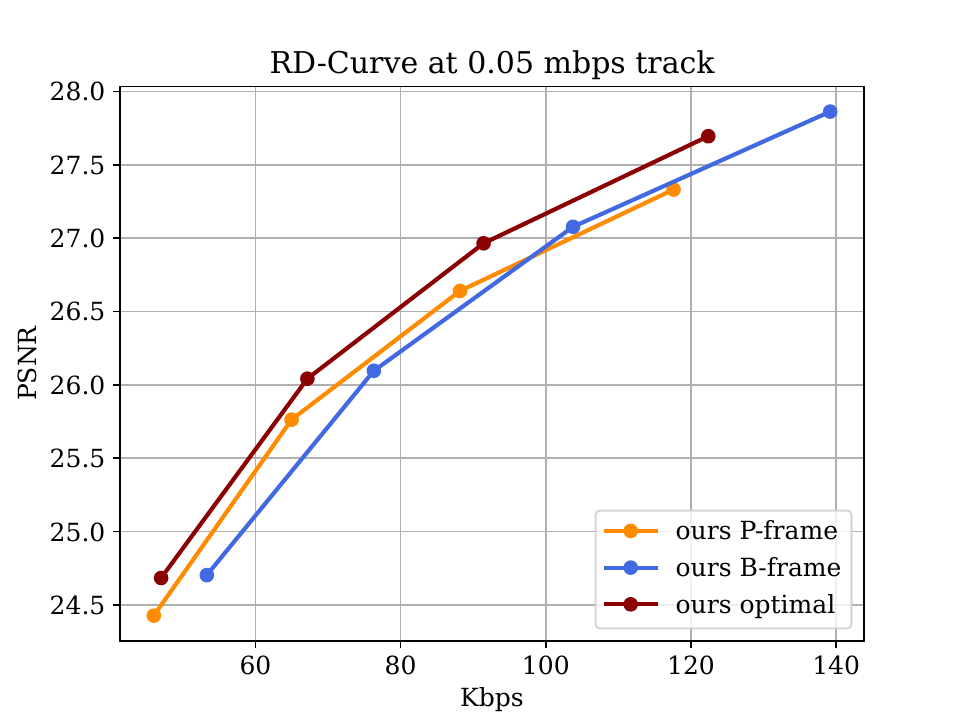}
%\caption{fig1}
}
\subfigure{
\includegraphics[width=0.45\linewidth]{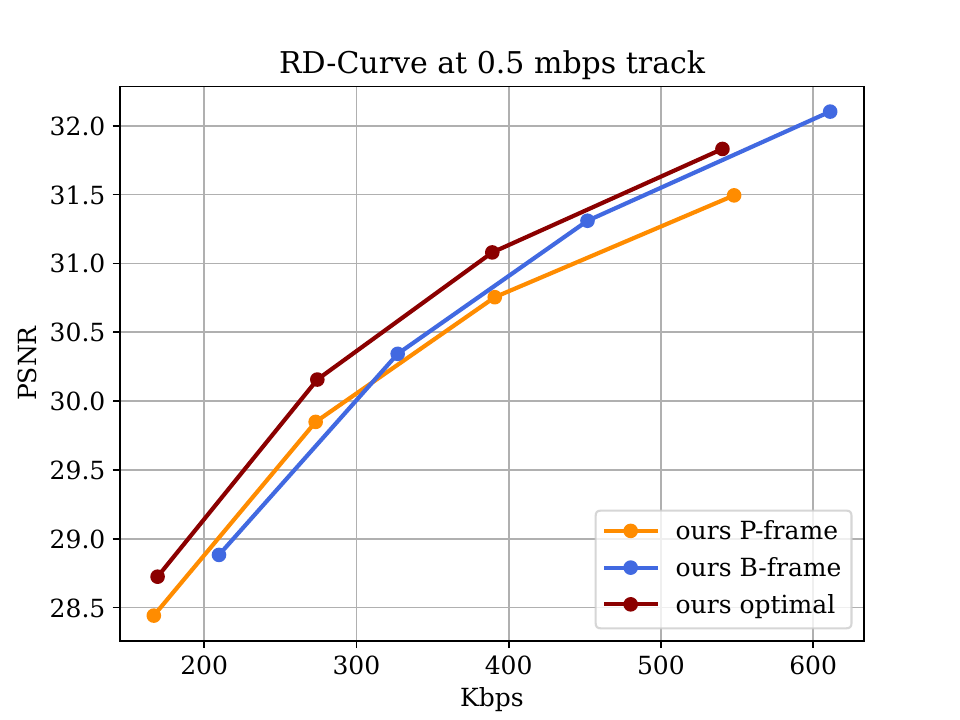}
}
\caption{RD-Curve in terms of PSNR on CLIC validation set at 0.05 mbps and 0.5 mbps. Results of different frame types are reported, where \textit{optimal} represents selecting the frame type that achieves better compression efficiency for each sequence.}
\label{RD_curve_clic}
\end{figure*}

\begin{figure*}[]
\centering
\includegraphics[width=\linewidth]{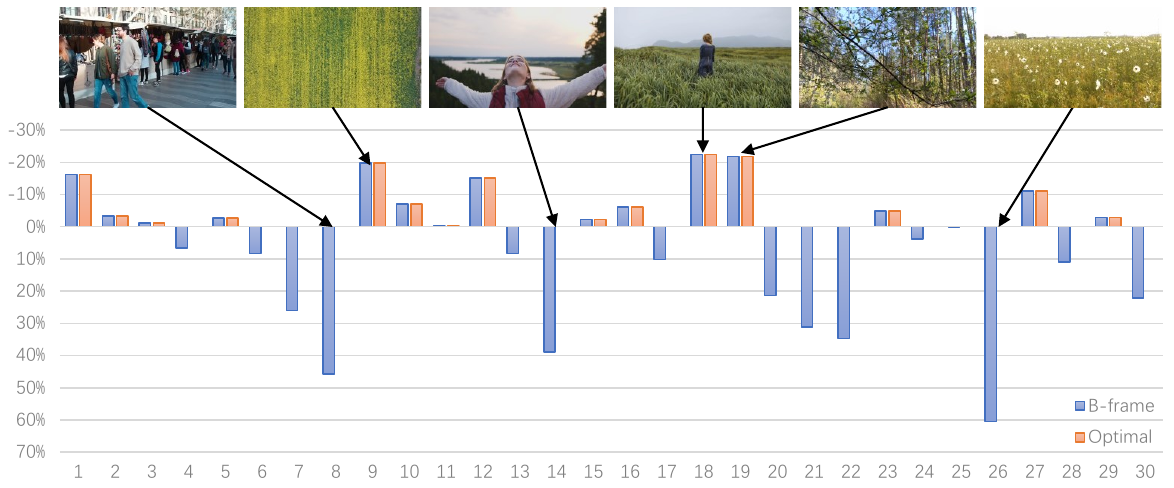} % Reduce the figure size so that it is slightly narrower than the column.
\caption{BD-Rate comparison in terms of PSNR on CLIC validation dataset at 0.05 mbps track. We set P-frame as anchor and report the result of B-frame and optimal frame type. The horizontal coordinates indicate the sequence index. We present video contents of several sequences, in which sequences 9, 18, 19 with slow and simple motion achieve bit rate savings on B-frame compression while sequences 8, 14, 26 with large motion, occlusion and camera motion suffer bit rate increments. By combining the optimal results of P-frame and B-frame coding, the framework can achieve better compression efficiency.
}
\label{bdbr_clic}
\end{figure*}

\subsection{Evaluation Results on CLIC Validation Dataset}
We report the rate-distortion performance of our framework on CLIC validation set by RD curves, in which Kbps and PSNR are averaged on all sequences. The GoP Size is set as 32 and all frames are compressed. 
The results of different frame types are summarized in Fig. \ref{RD_curve_clic}, where \textit{optimal} denotes combining the optimal results of P-frame and B-frame.
Besides, we present the per-sequence BD-Rate \cite{bjontegaard2001calculation} at 0.05 mbps track in Fig. \ref{bdbr_clic} for better illustration. We set P-frame as anchor and report the result of B-frame and optimal frame type. The negative number of BD-Rate denotes bitrate savings and positive number means bitrate increments.
It is shown that compressing B-frame brings efficiency loss on sequences 4, 6, 7, 8, 13, 14, 17, 20, 21, 22, 24, 25, 26, 28 and 30. As for the other sequences, B-frame compression improves the performance. 
The results demonstrate that sequences with large and complicated motions suffer performance loss on B-frame compression, while sequences with slow and simple motions can benefit from B-frame compression.
The reason is that in the hierarchical bi-directional prediction structure of B-frame compression, the current frame may have a large frame count interval from the reference frames, which reduces prediction performance in large and complicated motion scenes due to the low content similarity. As for slow and simple motion scenes, bi-directional prediction improves prediction performance due to the high content similarity.
By selecting appropriate frame types for each sequence, the optimal compression efficiency can be achieved. Note that an additional flag that indicates frame type needs to be transmitted to decoder side to reconstruct the sequence properly. 

\begin{figure*}[t]
\centering
\subfigure{
\includegraphics[width=0.3\linewidth]{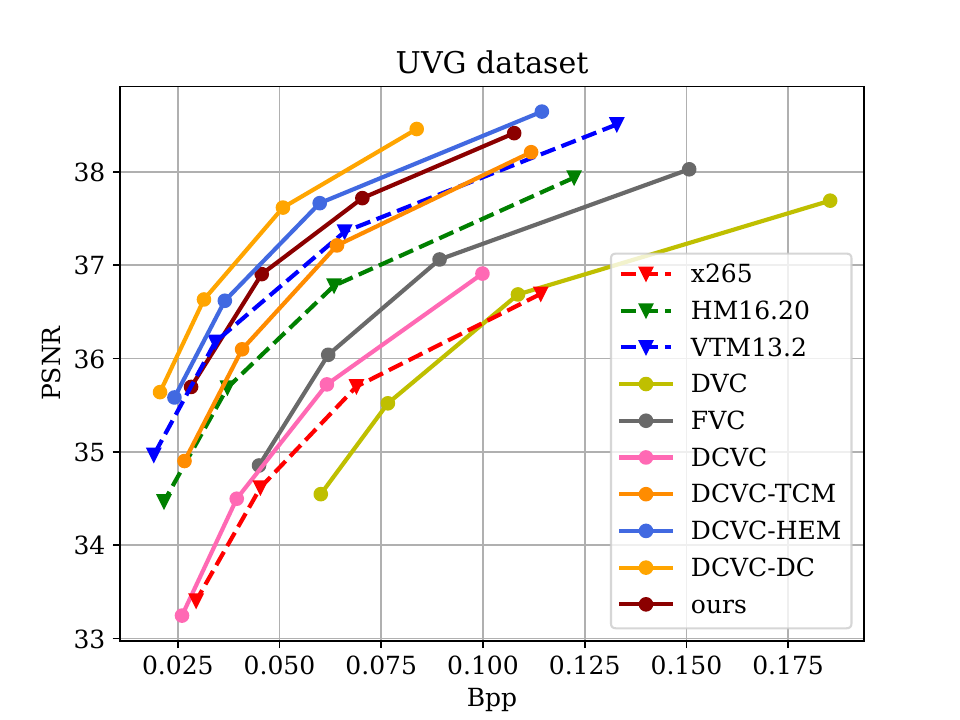}
%\caption{fig1}
}
\subfigure{
\includegraphics[width=0.3\linewidth]{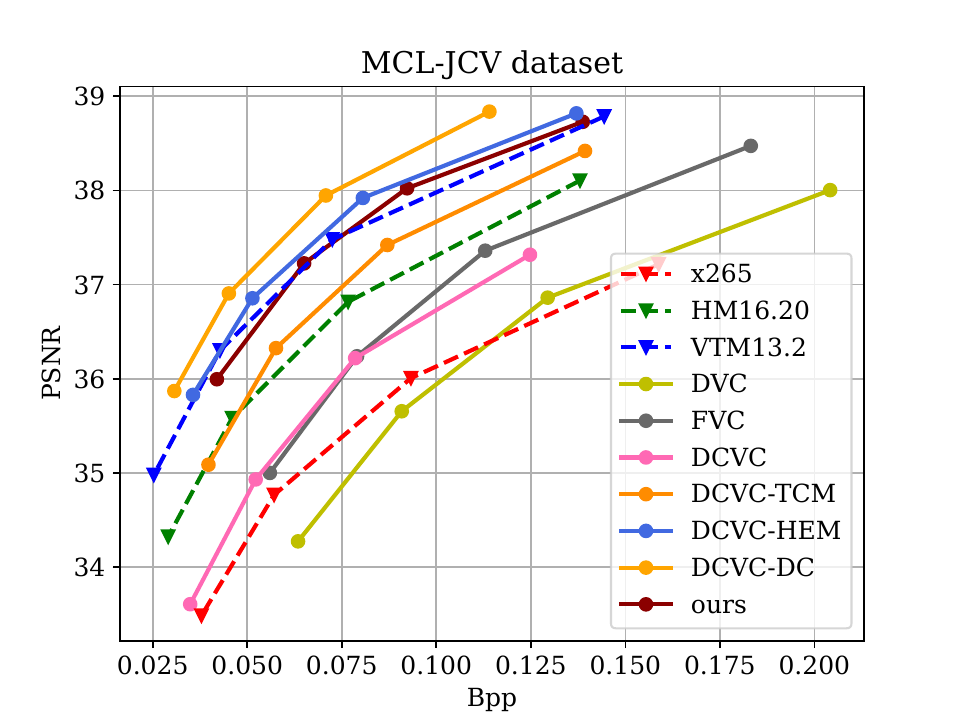}
}
\subfigure{
\includegraphics[width=0.3\linewidth]{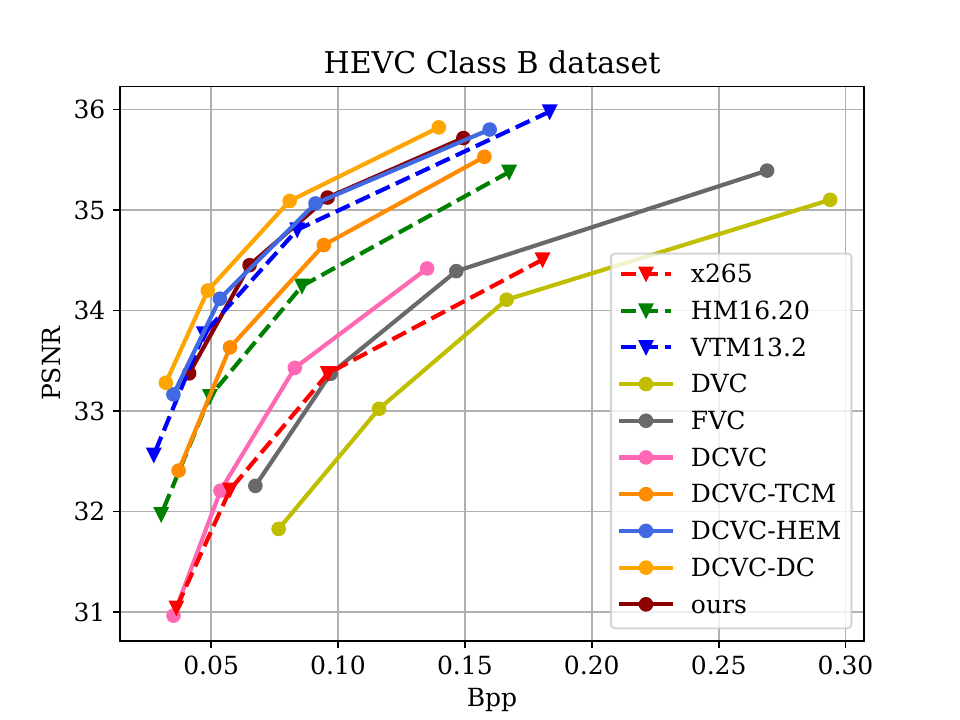}
}
\\
\subfigure{
\includegraphics[width=0.3\linewidth]{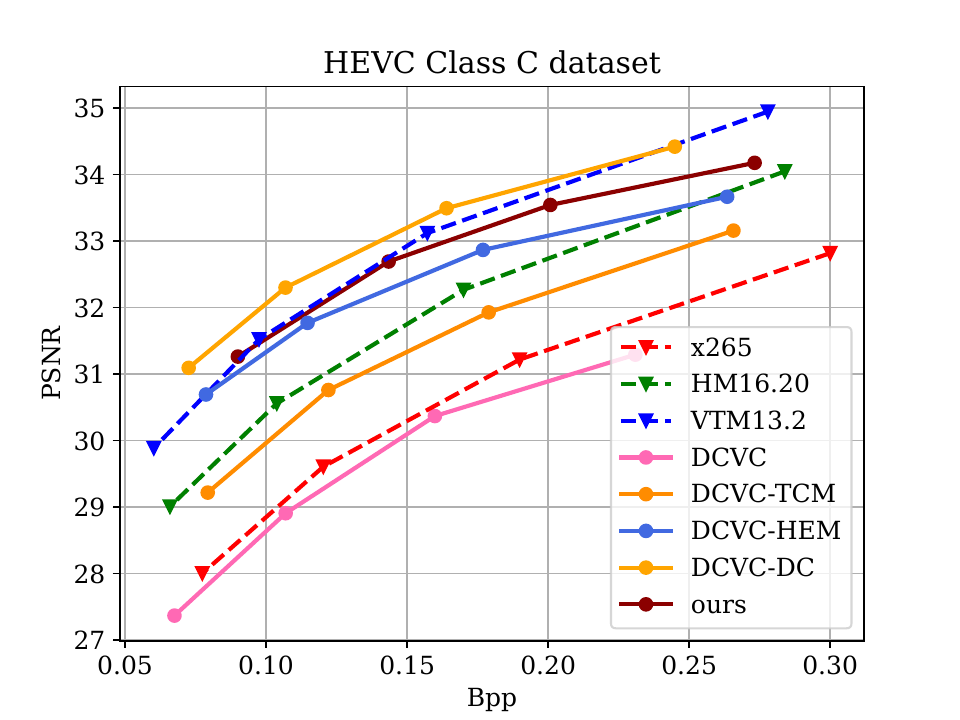}
}
\subfigure{
\includegraphics[width=0.3\linewidth]{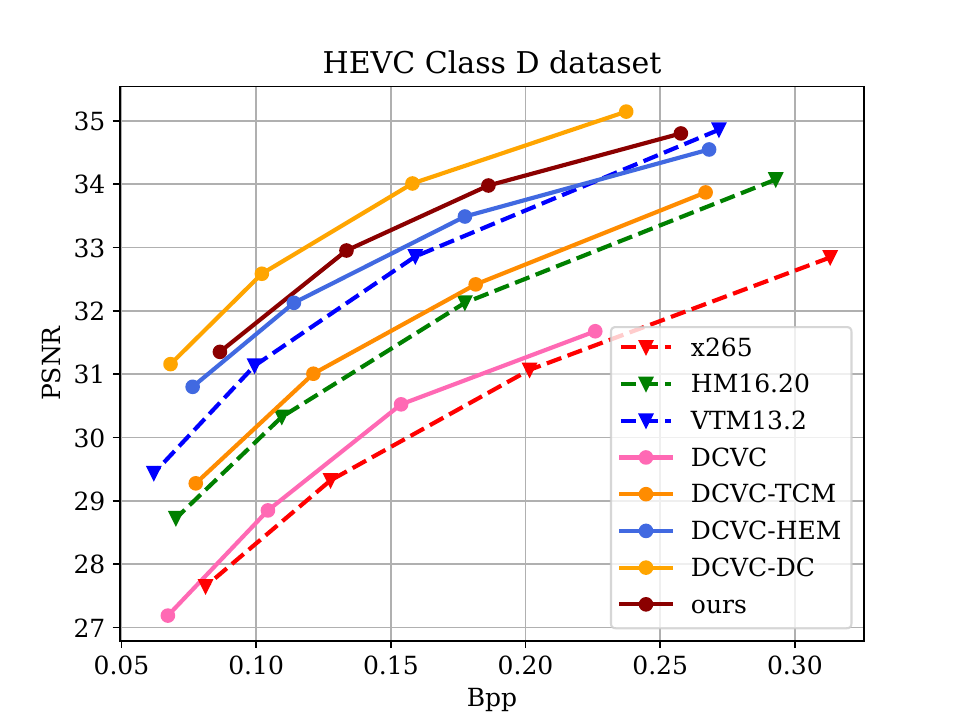}
}
\subfigure{
\includegraphics[width=0.3\linewidth]{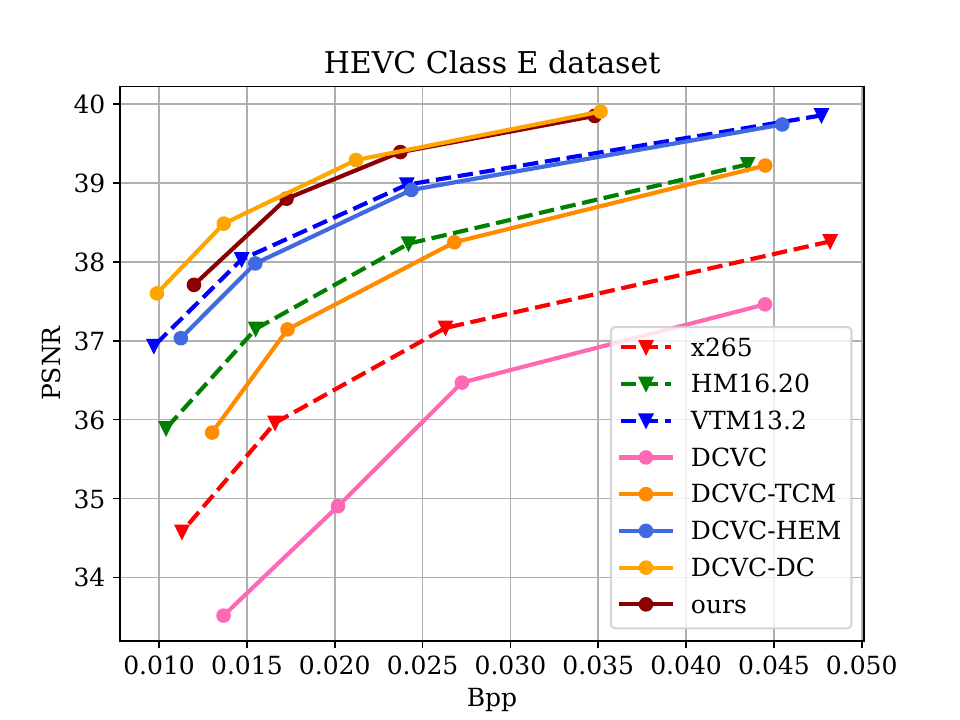}
}
\caption{Rate-distortion performance in terms of PSNR compared with traditional standards and recent learned methods on benchmark datasets.}
\label{RD_curve_g32_PSNR}
\end{figure*}

\begin{figure*}[t]
\centering
\subfigure{
\includegraphics[width=0.3\linewidth]{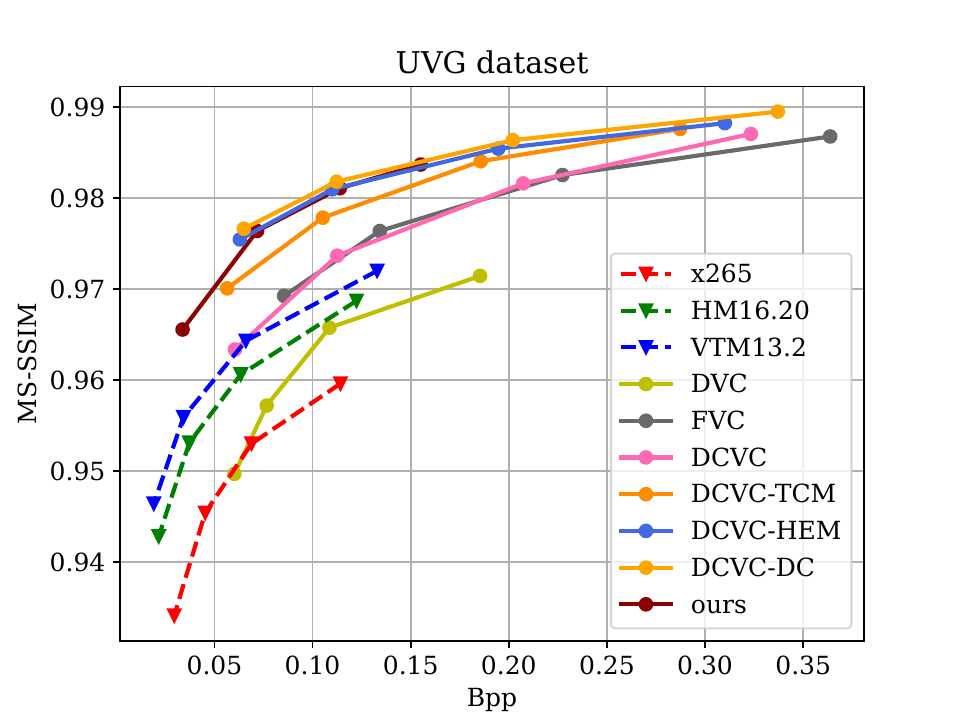}
%\caption{fig1}
}
\subfigure{
\includegraphics[width=0.3\linewidth]{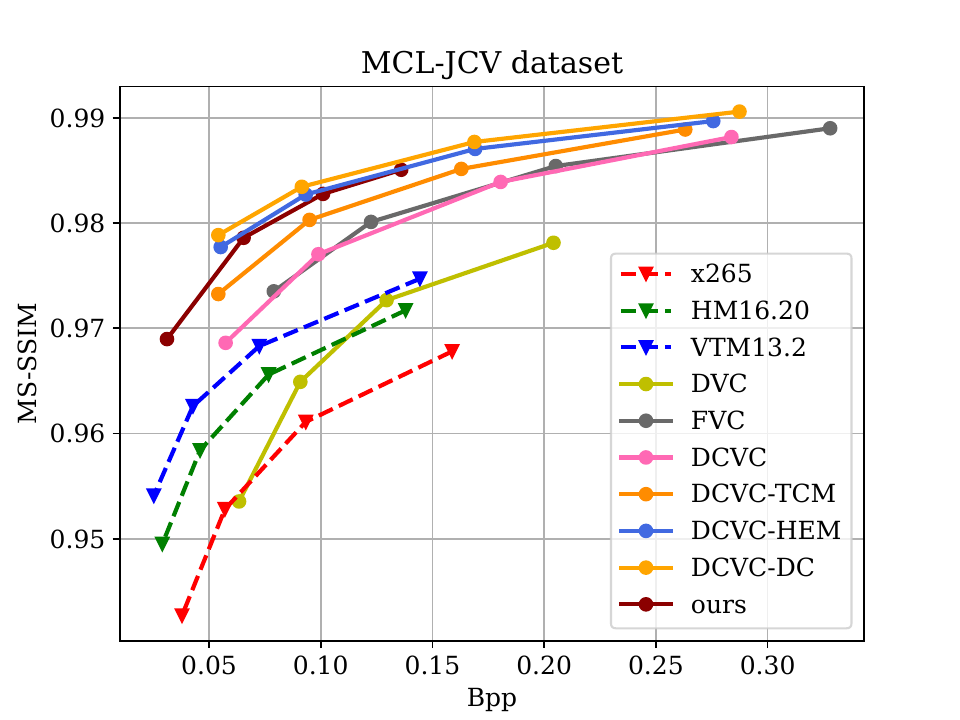}
}
\subfigure{
\includegraphics[width=0.3\linewidth]{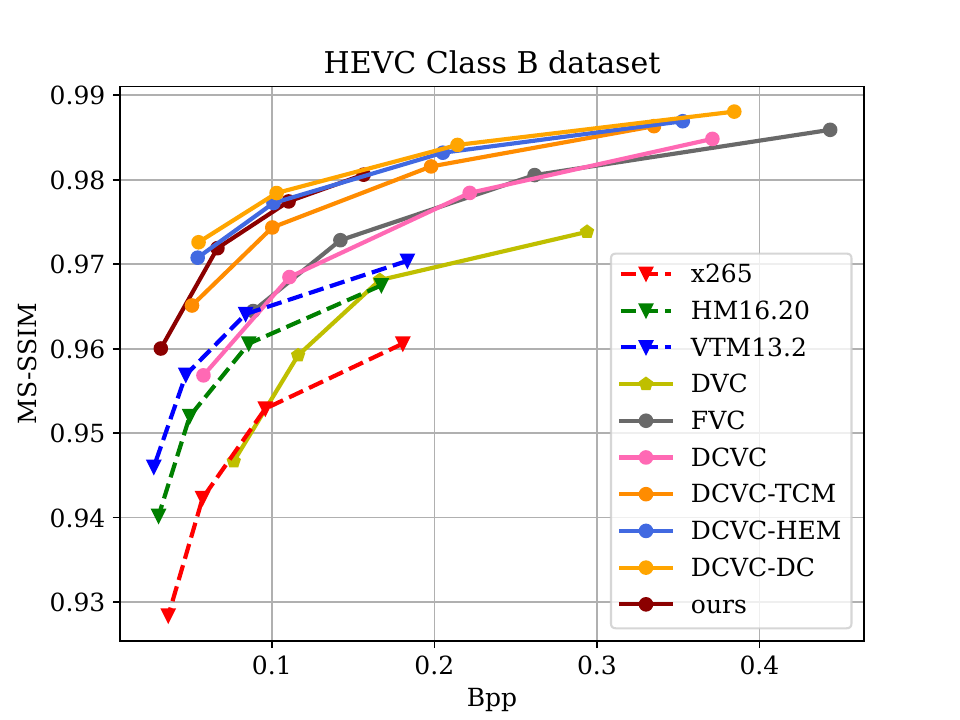}
}
\\
\subfigure{
\includegraphics[width=0.3\linewidth]{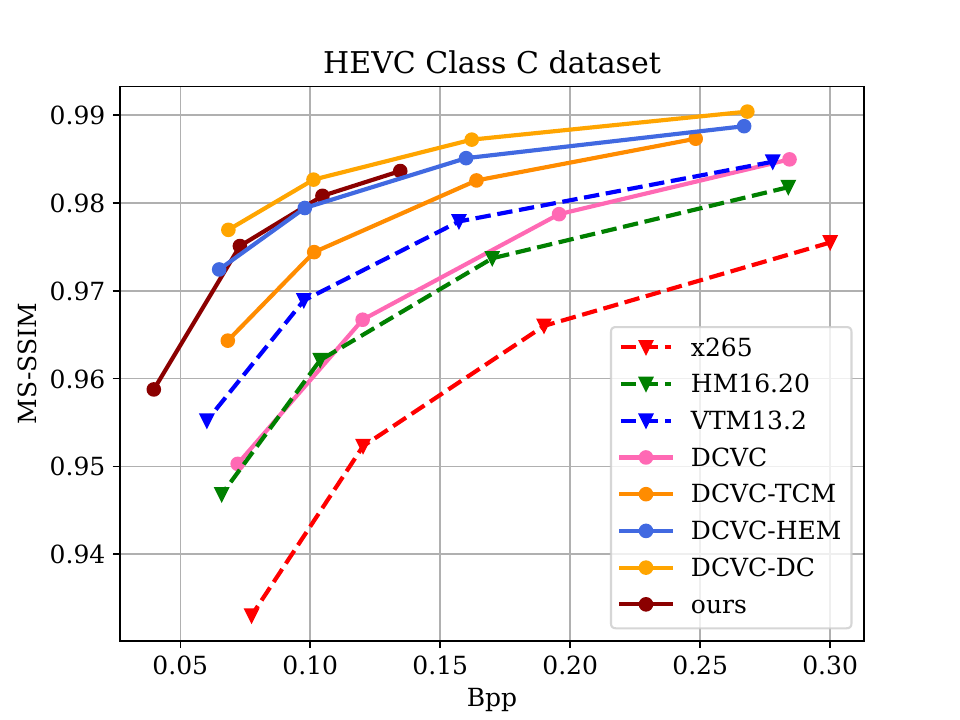}
}
\subfigure{
\includegraphics[width=0.3\linewidth]{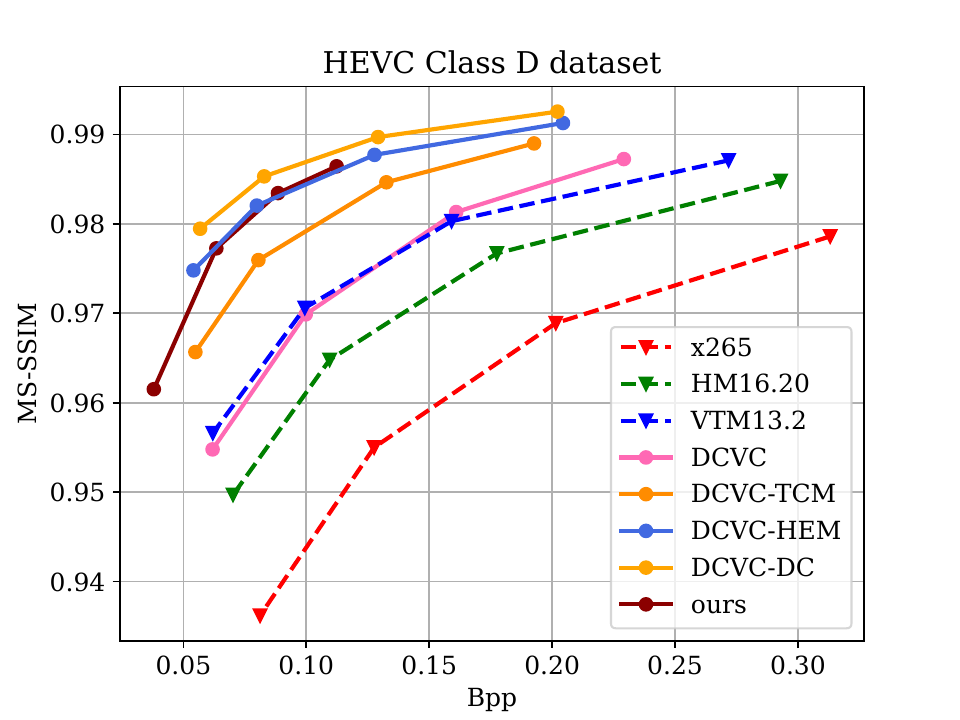}
}
\subfigure{
\includegraphics[width=0.3\linewidth]{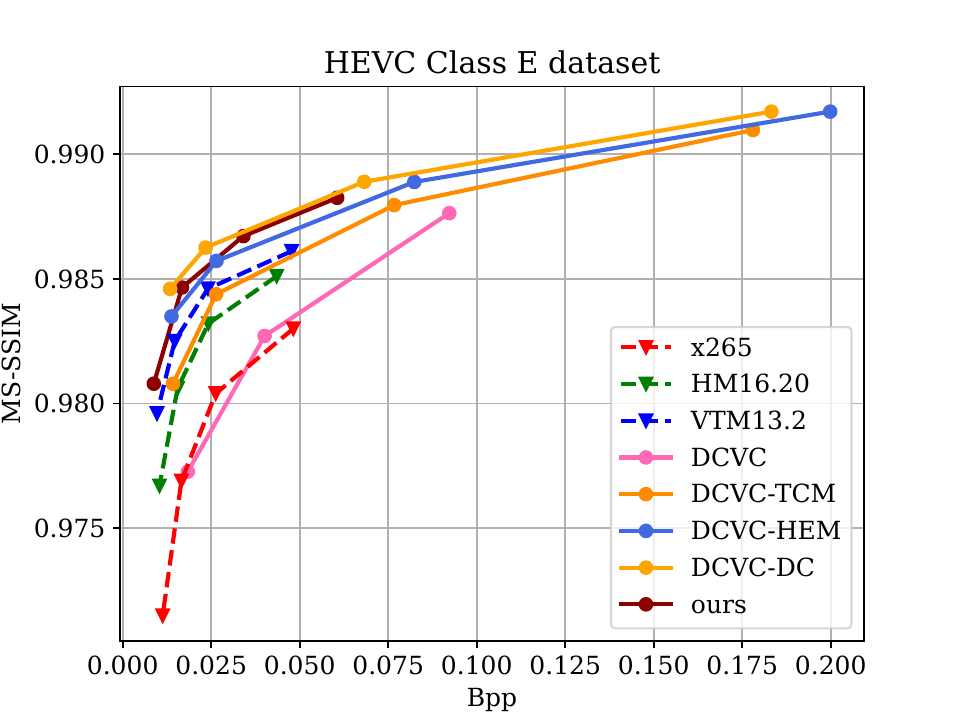}
}
\caption{Rate-distortion performance in terms of MS-SSIM compared with traditional standards and recent learned methods on benchmark datasets.}
\label{RD_curve_g32_msssim}
\end{figure*}

\begin{table*}[]
\footnotesize
% \scriptsize
\centering
\caption{BD-Rate comparison on PSNR with VTM as anchor.}
\begin{tabular}{c|ccc|cccc|c}
\toprule[1pt]
             & VTM & x265  & HM   & DCVC  & DCVC-TCM  & DCVC-HEM & DCVC-DC  & ours  \\ \hline
UVG\_1080P   & 0.0 & 157.1 & 35.7 & 120.5 & 16.7 & -18.1    & -30.0      & -9.0   \\
MCL-JCV      & 0.0 & 146.1 & 43.2 & 93.4 & 30.5 & -6.8      & -20.1      & 1.5  \\
HEVC B & 0.0 & 149.0 & 41.0 & 109.1 & 28.0 & -5.2     & -17.4      & -2.8   \\
HEVC C & 0.0 & 112.7 & 39.7 & 136.2 & 60.8 & 15.2     & -9.7       & 6.3  \\
HEVC D & 0.0 & 105.6 & 36.2 & 87.1 & 27.8 & -8.9     & -29.2       & -15.8 \\
HEVC E & 0.0 & 166.9 & 50.3 & 258.9 & 64.2 & 7.2      & -26.0      & -19.7 \\ \hline
Average      & 0.0 & 139.6 & 41.0 & 134.2 & 38.0 & -2.8      & -22.1     & -6.6   \\ \bottomrule[1pt]

\end{tabular}
\label{bdbr_psnr}
\end{table*}

\begin{table*}[]
\footnotesize
\centering
\caption{BD-Rate comparison on MS-SSIM with DCVC-TCM as anchor.}
\begin{tabular}{c|cccc|c}
\toprule[1pt]
             & DCVC-TCM & DCVC & DCVC-HEM & DCVC-DC  & ours  \\ \hline
UVG\_1080P   & 0.0 & 43.7 & -18.3    & -24.1        & -20.8 \\
MCL-JCV      & 0.0 & 32.8 & -22.2    & -30.1        & -19.3 \\
HEVC Class B & 0.0 & 59.0 & -19.3    & -26.5        & -17.7 \\
HEVC Class C & 0.0 & 56.6 & -25.4    & -38.6        & -30.1 \\
HEVC Class D & 0.0 & 53.9 & -27.9    & -40.6        & -25.6 \\
HEVC Class E & 0.0 & 60.0 & -24.6    & -39.0        & -38.0 \\ \hline
Average      & 0.0 & 51.0 & -23.0    & -33.2        & -25.3 \\ \bottomrule[1pt]
\end{tabular}
\label{bdbr_ssim}
\end{table*}

\subsection{Evaluation Results on Benchmark Datasets}
Following the settings of DCVC-TCM, we set GoP size as 32 and encode 96 frames for all test sequences. 
We report the optimal results of our method by selecting appropriate frame type between P-frame and B-frame for each sequence, where P-frame achieves better compression efficiency on sequences with large and complicated motions while B-frame performs better on sequences with slow and simple motions.
For comparison with traditional methods, we evaluate the compression efficiency of the test model implementation of HEVC and VVC, referred to as HM and VTM, and open source library implementation of HEVC, referred to as x265. For x265, we test the implementation of FFmpeg under \textit{veryslow} preset. For reference software, we test \textit{encoder\_lowdelay\_main.cfg} configuration of HM-16.20 and \textit{encoder\_lowdelay\_vtm.cfg} configuration of VTM-13.2.  
The test sequences are converted from YUV420 to YUV444 for traditional codecs, then the distortion are measured in RGB444 space.
For comparison with existing learned methods, we test the representative methods DVC \cite{lu2019dvc}, FVC \cite{hu2021fvc}, DCVC \cite{li2021deep}, DCVC-TCM \cite{sheng2022temporal}, DCVC-HEM \cite{li2022hybrid} and DCVC-DC \cite{li2023neural}.

We present rate-distortion performance on PSNR and MS-SSIM by RD curves and BD-Rate. Bit per pixel (Bpp) and PSNR of RD curves are averaged on all sequences within each testing dataset. And BD-Rate is calculated on average Bpp and PSNR.
In terms of PSNR, the results on benchmark datasets are summarized in Fig. \ref{RD_curve_g32_PSNR} and Table \ref{bdbr_psnr}. We use VTM as anchor to calculate BD-Rate. It is shown that our method outperforms traditional methods x265, HM, learned method DCVC-TCM, and achieves comparable performance with VTM, DCVC-HEM. Specifically, our method with \textit{optimal} setting presents $6.6\%$ bit rate savings over VTM averaged on all test datasets.
In terms of MS-SSIM, the results are shown in Fig. \ref{RD_curve_g32_msssim} and Table \ref{bdbr_ssim}. Due to the misalignment of bit rate between traditional methods and learned methods, we switch to DCVC-TCM as anchor to calculate BD-Rate. The results show that our method outperforms DCVC, DCVC-TCM and is comparable with DCVC-HEM. As for comparison with traditional methods, it can be concluded from RD curves that learned methods outperform traditional methods under most test conditions, since learned methods can be optimized by metric-oriented distortion loss.

% \begin{table*}[]
% \scriptsize
% \centering
% \caption{Comparison on encoding and decoding time of learned and non-learned methods.}
% \begin{tabular}{c|ccc|cc|cc}
% \toprule[1pt]
% Time (s) & x265 & HM & VTM & DCVC-TCM*  & ours P/B-frame* & ours P/B-frame & ours optimal  \\ \hline
% Encoding & 1.962 & 22.135 & 99.792  & 0.557   & 0.641     & 0.997    &  1.995   \\
% Decoding & 0.008 & 0.057  & 0.075   & 0.164   & 0.211     & 0.555    &  0.555   \\ 
% \bottomrule[1pt]
% \end{tabular}
% \label{speed}
% \\
% \textit{method*} denotes model inference time without entropy coding or decoding.
% \end{table*}

% \subsection{Model Complexity} 
% We compare the complexity of by encoding and decoding time of a 1080P-frame. The results are summarized in Table \ref{speed}. For learned methods, we only report the model inference time due to different entropy encoder and decoder are used, which are denoted as \textit{method*}. For non-learned methods, we report the overall encoding and decoding time, which are denoted as \textit{method}. The results show that learned methods have faster encoding speed but slower decoding speed than non-learned methods. 
% As for comparison with DCVC-TCM, our method shows slower speed. 
% The reason is that our framework involves two reference frames, which introduces additional computation costs.
% When enabling the \textit{optimal} frame type setting, the encoding time will be doubled than that of P/B-frame since both P-frame and B-frame compression needs to be conducted. Nonetheless, the decoding time remains unchanged because only one-pass decoding is performed.

\section{Conclusion}

In this paper, we present a unified contextual video compression (UCVC) with joint P-frame and B-frame coding in response to video compression track of the 6th Challenge on Learned Image Compression (CLIC) at DCC 2024.
Our framework takes two decoded frames as references, where the frames can be either both from the past for P-frame compression or one from the past and one from the future for B-frame compression.
Based on the flexibility, we report the optimal compression efficiency by selecting appropriate frame type for each sequence.

\Section{References}
\small
\bibliographystyle{IEEEbib}
\bibliography{refs}

\begin{thebibliography}{10}

\bibitem{wiegand2003overview}
Thomas Wiegand, Gary~J Sullivan, Gisle Bjontegaard, and Ajay Luthra,
\newblock ``Overview of the h. 264/avc video coding standard,''
\newblock {\em IEEE Transactions on circuits and systems for video technology}, vol. 13, no. 7, pp. 560--576, 2003.

\bibitem{sullivan2012overview}
Gary~J Sullivan, Jens-Rainer Ohm, Woo-Jin Han, and Thomas Wiegand,
\newblock ``Overview of the high efficiency video coding (hevc) standard,''
\newblock {\em IEEE Transactions on circuits and systems for video technology}, vol. 22, no. 12, pp. 1649--1668, 2012.

\bibitem{bross2021overview}
Benjamin Bross, Ye-Kui Wang, Yan Ye, Shan Liu, Jianle Chen, Gary~J Sullivan, and Jens-Rainer Ohm,
\newblock ``Overview of the versatile video coding (vvc) standard and its applications,''
\newblock {\em IEEE Transactions on Circuits and Systems for Video Technology}, vol. 31, no. 10, pp. 3736--3764, 2021.

\bibitem{lu2019dvc}
Guo Lu, Wanli Ouyang, Dong Xu, Xiaoyun Zhang, Chunlei Cai, and Zhiyong Gao,
\newblock ``Dvc: An end-to-end deep video compression framework,''
\newblock in {\em Proceedings of the IEEE/CVF Conference on Computer Vision and Pattern Recognition}, 2019, pp. 11006--11015.

\bibitem{hu2021fvc}
Zhihao Hu, Guo Lu, and Dong Xu,
\newblock ``Fvc: A new framework towards deep video compression in feature space,''
\newblock in {\em Proceedings of the IEEE/CVF Conference on Computer Vision and Pattern Recognition}, 2021, pp. 1502--1511.

\bibitem{li2021deep}
Jiahao Li, Bin Li, and Yan Lu,
\newblock ``Deep contextual video compression,''
\newblock {\em Advances in Neural Information Processing Systems}, vol. 34, 2021.

\bibitem{sheng2022temporal}
Xihua Sheng, Jiahao Li, Bin Li, Li~Li, Dong Liu, and Yan Lu,
\newblock ``Temporal context mining for learned video compression,''
\newblock {\em IEEE Transactions on Multimedia}, 2022.

\bibitem{li2022hybrid}
Jiahao Li, Bin Li, and Yan Lu,
\newblock ``Hybrid spatial-temporal entropy modelling for neural video compression,''
\newblock in {\em Proceedings of the 30th ACM International Conference on Multimedia}, 2022, pp. 1503--1511.

\bibitem{li2023neural}
Jiahao Li, Bin Li, and Yan Lu,
\newblock ``Neural video compression with diverse contexts,''
\newblock in {\em Proceedings of the IEEE/CVF Conference on Computer Vision and Pattern Recognition}, 2023, pp. 22616--22626.

\bibitem{wu2018video}
Chao-Yuan Wu, Nayan Singhal, and Philipp Krahenbuhl,
\newblock ``Video compression through image interpolation,''
\newblock in {\em Proceedings of the European conference on computer vision (ECCV)}, 2018, pp. 416--431.

\bibitem{djelouah2019neural}
Abdelaziz Djelouah, Joaquim Campos, Simone Schaub-Meyer, and Christopher Schroers,
\newblock ``Neural inter-frame compression for video coding,''
\newblock in {\em Proceedings of the IEEE/CVF International Conference on Computer Vision}, 2019, pp. 6421--6429.

\bibitem{yilmaz2021end}
M~Ak{\i}n Y{\i}lmaz and A~Murat Tekalp,
\newblock ``End-to-end rate-distortion optimized learned hierarchical bi-directional video compression,''
\newblock {\em IEEE Transactions on Image Processing}, vol. 31, pp. 974--983, 2021.

\bibitem{akin2023multi}
M~Ak{\i}n~Y{\i}lmaz, O~Ugur~Ulas, and A~Murat~Tekalp,
\newblock ``Multi-scale deformable alignment and content-adaptive inference for flexible-rate bi-directional video compression,''
\newblock 2023.

\bibitem{cheng2020image}
Zhengxue Cheng, Heming Sun, Masaru Takeuchi, and Jiro Katto,
\newblock ``Learned image compression with discretized gaussian mixture likelihoods and attention modules,''
\newblock in {\em Proceedings of the IEEE Conference on Computer Vision and Pattern Recognition (CVPR)}, 2020.

\bibitem{Cui_2021_CVPR}
Ze~Cui, Jing Wang, Shangyin Gao, Tiansheng Guo, Yihui Feng, and Bo~Bai,
\newblock ``Asymmetric gained deep image compression with continuous rate adaptation,''
\newblock in {\em Proceedings of the IEEE/CVF Conference on Computer Vision and Pattern Recognition (CVPR)}, June 2021, pp. 10532--10541.

\bibitem{minnen2018joint}
David Minnen, Johannes Ball{\'e}, and George~D Toderici,
\newblock ``Joint autoregressive and hierarchical priors for learned image compression,''
\newblock {\em Advances in neural information processing systems}, vol. 31, 2018.

\bibitem{ranjan2017optical}
Anurag Ranjan and Michael~J Black,
\newblock ``Optical flow estimation using a spatial pyramid network,''
\newblock in {\em Proceedings of the IEEE conference on computer vision and pattern recognition}, 2017, pp. 4161--4170.

\bibitem{xue2019video}
Tianfan Xue, Baian Chen, Jiajun Wu, Donglai Wei, and William~T Freeman,
\newblock ``Video enhancement with task-oriented flow,''
\newblock {\em International Journal of Computer Vision}, vol. 127, no. 8, pp. 1106--1125, 2019.

\bibitem{mercat2020uvg}
Alexandre Mercat, Marko Viitanen, and Jarno Vanne,
\newblock ``Uvg dataset: 50/120fps 4k sequences for video codec analysis and development,''
\newblock in {\em Proceedings of the 11th ACM Multimedia Systems Conference}, 2020, pp. 297--302.

\bibitem{wang2016mcl}
Haiqiang Wang, Weihao Gan, Sudeng Hu, Joe~Yuchieh Lin, Lina Jin, Longguang Song, Ping Wang, Ioannis Katsavounidis, Anne Aaron, and C-C~Jay Kuo,
\newblock ``Mcl-jcv: a jnd-based h. 264/avc video quality assessment dataset,''
\newblock in {\em 2016 IEEE international conference on image processing (ICIP)}. IEEE, 2016, pp. 1509--1513.

\bibitem{begaint2020compressai}
Jean B{\'e}gaint, Fabien Racap{\'e}, Simon Feltman, and Akshay Pushparaja,
\newblock ``Compressai: a pytorch library and evaluation platform for end-to-end compression research,''
\newblock {\em arXiv preprint arXiv:2011.03029}, 2020.

\bibitem{loshchilov2018decoupled}
Ilya Loshchilov and Frank Hutter,
\newblock ``Decoupled weight decay regularization,''
\newblock in {\em International Conference on Learning Representations}, 2018.

\bibitem{bjontegaard2001calculation}
Gisle Bjontegaard,
\newblock ``Calculation of average psnr differences between rd-curves,''
\newblock {\em VCEG-M33}, 2001.

\end{thebibliography}

\end{document}